\documentclass[conference]{IEEEtran}

\IEEEoverridecommandlockouts

\usepackage{cite}

\usepackage{amsmath,amssymb,amsfonts}

\usepackage{algorithmic}

\usepackage{graphicx}

\usepackage{textcomp}

\usepackage{xcolor}

\usepackage{enumitem}

\usepackage{booktabs}

\def\BibTeX{{\rm B\kern-.05em{\sc i\kern-.025em b}\kern-.08em

\kern-.1667em\lower.7ex\hbox{\rm X}}}

\begin{document}

\title{Single-Pass Uncertainty Heads for Claim-Level Hallucination Detection in Persian Medical Language Models}

\author{\IEEEauthorblockN{1\textsuperscript{st} Mehrdad Ghassabi}

\IEEEauthorblockA{\textit{Faculty of Computer Engineering} \\
\textit{University of Isfahan} \\
Isfahan, Iran \\
m.ghassabi@eng.ui.ac.ir}

\and

\IEEEauthorblockN{2\textsuperscript{nd} Pedram Rostami}

\IEEEauthorblockA{\textit{School of Electrical and Computer Engineering} \\
\textit{University of Tehran} \\
Tehran, Iran \\
pedram.rostami@ut.ac.ir}

\and

\IEEEauthorblockN{3\textsuperscript{rd} Hamidreza Baradaran Kashani}

\IEEEauthorblockA{\textit{Faculty of Computer Engineering} \\
\textit{University of Isfahan} \\
Isfahan, Iran \\
hrb.kashani@eng.ui.ac.ir}

\and

\IEEEauthorblockN{4\textsuperscript{th} Sadra Hakim}

\IEEEauthorblockA{\textit{School of Computer Science} \\
\textit{University of Windsor} \\
Windsor, Canada \\
hakim6@uwindsor.ca}

\and

\IEEEauthorblockN{5\textsuperscript{th} Audrina Ebrahimi}

\IEEEauthorblockA{\textit{Center for Brain Health} \\
\textit{University of Texas at Dallas} \\
TX, US \\
audrina.ebrahimi@utdallas.edu}

}

\maketitle

\begin{abstract}

Hallucination detection is particularly important for medical language models, but repeated-sampling approaches are computationally expensive. A faster alternative is a single-pass uncertainty head that predicts hallucination risk from a frozen generator's internal signals. Existing uncertainty heads consume backbone-specific features and tokenization, motivating adaptation when the backbone or language changes. We study two Persian medical models, Gaokerena-V and Gaokerena-R, on a 168-question Iranian medical entrance examination. Across five generations per question, Gaokerena-V produces the same option on only 14 questions and Gaokerena-R on 37, compared with 168 for Med-Gemma, indicating substantial response variability in the Gaokerena models. We therefore adapt the LLM Uncertainty Head (LUH) framework to these models and construct two paired claim-level hallucination datasets directly in Persian, with 1,600 responses per backbone. Lightweight heads are trained on frozen-backbone attention maps and token probabilities. On held-out test splits, the heads achieve PR-AUCs of 0.4820 and 0.4652, corresponding to 2.30 and 2.66 times their respective random baselines, and ROC-AUCs of 0.7852 and 0.7810. The resulting detectors require neither retrieval nor repeated sampling at inference.

\end{abstract}


\section{Introduction}

Large language models are increasingly used in settings where unreliable outputs can be costly, making uncertainty estimation and hallucination detection important. This challenge is especially relevant for low-resource languages, where language-specific evaluation resources and tools remain limited. In medical question answering, even a strong generator can produce individual unsupported claims, so a detector that identifies claims requiring verification can provide a useful safety signal without modifying the generator itself.

A common approach to uncertainty estimation is to generate multiple responses and measure their agreement. Consistency-based methods can be informative, but repeated sampling increases inference cost and generally produces a response-level signal rather than a direct claim-level warning. White-box models offer another possibility: internal signals such as attention patterns and token probabilities can be processed by an auxiliary head to estimate uncertainty while preserving the original generator and avoiding multiple generations.

In this work, we adapt the LLM Uncertainty Head (LUH) framework of Shelmanov et al.~\cite{b1} to the Aya-Expanse-8B backbone and Persian medical generation. We study two previously developed Persian medical models, Gaokerena-V and Gaokerena-R~\cite{b2,b3}, but the present work focuses on the uncertainty-estimation layer rather than on the construction of those models. Their medical adaptation and reasoning-data construction are described in our earlier work.

Our study addresses three practical questions. First, do the Aya-Expanse-derived models exhibit sufficient response variability to motivate an uncertainty estimator beyond repeated-sampling consistency? Second, can LUH-style claim-level supervision be constructed directly in Persian rather than translated from an English resource? Third, can lightweight heads learn a useful hallucination signal from backbone-internal features for two related but differently trained Persian medical backbones?

Our contributions are:

\begin{enumerate}[leftmargin=*]

\item We characterize response variability of Aya-Expanse-8B, Gaokerena-V, and Gaokerena-R on a fixed 168-question Iranian medical entrance examination, using explicitly defined accuracy, consistency, and entropy measures, which motivates a single-pass alternative to repeated-sampling consistency estimates.

\item We construct two paired claim-level hallucination datasets directly in Persian, with 1,600 responses per backbone, and develop the corresponding token-level supervision for Aya-Expanse-8B tokenization.

\item We train and evaluate lightweight claim-level uncertainty heads for Gaokerena-V and Gaokerena-R. The heads obtain ROC-AUCs of 0.7852 and 0.7810 and PR-AUCs that are 2.30 and 2.66 times their respective random baselines.

\end{enumerate}

\section{Background and Related Work}

\subsection{Uncertainty and Hallucination Detection}

Uncertainty estimation methods for language models include verbal, latent-information, consistency-based, and semantic-clustering approaches~\cite{b4}. Verbal methods ask a model to report its own confidence; latent-information methods use internal representations or other signals available during generation; consistency-based methods estimate reliability from agreement across multiple generations; and semantic-clustering methods aggregate multiple outputs by meaning. For white-box models, latent-information approaches can provide uncertainty signals without requiring additional generations.

Uncertainty is also relevant to hallucination detection because hallucinated content can be associated with weaker or less stable model signals. In a medical setting, a detector does not necessarily have to replace the generator: it can instead identify claims that deserve additional verification~\cite{b5,b6}. This motivates a lightweight auxiliary model whose output is interpreted as a claim-level hallucination score.

Recent work has also explored uncertainty heads for claim or reasoning-step verification~\cite{b7}. These approaches separate generation from verification: the language model remains fixed while a smaller head learns to map internal features to a reliability estimate. Our work follows this general design but changes both the backbone and the language, requiring new heads trained from Persian supervision.

\subsection{Aya-Expanse and Gaokerena}

Aya-Expanse is a multilingual family designed to improve performance in underrepresented languages through large-scale synthetic data and iterative preference training~\cite{b8,b9,b10}. The two backbones studied here, Gaokerena-V and Gaokerena-R, are Persian medical models developed in our prior work~\cite{b2,b3}. Gaokerena-V is a medically adapted Aya-Expanse model, while Gaokerena-R is a reasoning-oriented variant. We use these models as fixed target backbones for the present uncertainty experiments and do not treat their original medical adaptation or reasoning-data construction as contributions of this paper.

\subsection{LLM Uncertainty Heads}

The LUH framework~\cite{b1} augments a pretrained language model with a lightweight auxiliary head that predicts whether extracted claims are hallucinated. The head uses internal backbone signals, including per-layer attention maps and token-probability statistics, while the backbone remains frozen. The original framework was developed for other model families and English data. Although released LUH heads can generalize in some cross-language settings, the head consumes backbone-specific internal features and tokenization, so an existing head is not directly interchangeable with a different architecture. We therefore train new heads for the Aya-Expanse-based models.

A second adaptation concerns supervision. Translating an existing English claim dataset into Persian would require translation followed by claim-span and token realignment. Because our goal is claim-level prediction tied to the exact generated response, we instead construct the responses, claims, labels, and token spans directly in Persian.

\section{Response Variability}

Before training the uncertainty heads, we examine whether repeated generations exhibit measurable variability. We evaluate Aya-Expanse-8B, Gaokerena-V, Gaokerena-R, and Med-Gemma~\cite{b11} on the September 2023 Iranian Basic Medical Sciences Entrance Examination (IBMSEE), containing 168 questions. Each model is run five times on the same questions using chain-of-thought prompting~\cite{b12}.

\subsection{Metrics}

For each question, the final option letter (A--D) is extracted from each of the five runs. A run for which no valid option letter can be extracted (for example, an output of ``none'' or ``multiple choice'') is treated as an invalid run; these account for 19 of 840 runs for Gaokerena-V, 35 of 840 for Gaokerena-R, and 19 of 840 for Aya-Expanse. We then compute four quantities.

\textit{Accuracy.} The ensemble answer of a question is the option chosen by at least three of the five runs; if no option reaches three votes, the question is marked as not agreed. Accuracy is the percentage of the 168 questions whose ensemble answer equals the correct option. Not-agreed questions are counted as incorrect.

\textit{Single option taken.} The number of questions for which all five runs yield the same valid option letter. A question containing an invalid run is therefore not counted.

\textit{All options taken.} The number of questions for which all four options (A, B, C, and D) appear among the valid answers of the five runs.

\textit{Entropy.} For question $q$, let $n_q$ be the number of valid runs and $p_{q,o}$ the fraction of those runs that select option $o$. The predictive entropy is

\[
H_q = -\sum_{o\in\{A,B,C,D\}} p_{q,o}\,\log_2 p_{q,o},
\]

with $0\log_2 0 = 0$, and the value reported in Table~\ref{tab:instability} is the mean of $H_q$ over the 168 questions, in bits. Invalid runs are excluded from the distribution, and a question with no valid run is assigned $H_q=0$. $H_q=0$ when all valid runs agree; with five runs and four options, the maximum is approximately $1.92$ bits (a 2--1--1--1 split).

\begin{table}[ht]

\centering

\small

\caption{Five-run consistency on the September 2023 IBMSEE. Accuracy is the ensemble-answer accuracy (\%) with not-agreed questions counted as incorrect; entropy is the mean per-question Shannon entropy in bits over valid runs (see text).}

\label{tab:instability}

\resizebox{\columnwidth}{!}{%

\begin{tabular}{|l|c|c|c|c|}

\hline

& \textbf{Gao} & \textbf{Gao} & \textbf{Aya-} & \textbf{Med-} \\

& \textbf{kerena-V} & \textbf{kerena-R} & \textbf{Expanse} & \textbf{Gemma} \\ \hline

Accuracy & 29.76 & 38.69 & 35.71 & 37.50 \\ \hline

Questions with a & 14 & 37 & 36 & 168 \\

single option taken &  &  &  &  \\ \hline

Questions with all  & 10 & 4 & 6 & 0 \\

 options taken &  &  &  & \\ \hline

Entropy (bits) & 1.11 & 0.80 & 0.84 & 0 \\ \hline

Prompt & CoT & CoT & CoT & CoT \\ \hline

Parameters & 8B & 8B & 8B & 4B \\ \hline

\end{tabular}}

\end{table}

\subsection{Findings}

Med-Gemma selected the same option on all 168 questions. In contrast, Gaokerena-V was fully consistent on only 14 questions, compared with 36 for Aya-Expanse and 37 for Gaokerena-R. Gaokerena-V also has higher entropy (1.11 bits) than Aya-Expanse (0.84) and Gaokerena-R (0.80), and the lowest ensemble accuracy (29.76\%), in part because no option reached a three-vote majority on 48 of its questions, compared with 28 for Aya-Expanse and 32 for Gaokerena-R. Thus, under this evaluation setup, Gaokerena-V exhibits substantially greater response variability than the original Aya-Expanse-8B model, whereas Gaokerena-R is comparable to Aya-Expanse-8B in consistency and entropy and attains the highest ensemble accuracy among the three 8B models. The number of questions on which all four options appear is small for every model (between 4 and 10), so we do not interpret differences in that row.

The table also illustrates why consistency-based uncertainty can be useful but costly. Five generations are required for every question, and the resulting statistics describe agreement over answer options rather than the reliability of individual claims within the generated explanation. A claim-level head can instead assign a score to each extracted claim from a single generation. We therefore use this variability experiment as motivation rather than evidence for a general effect of medical adaptation: the variability differs between the two Gaokerena variants, it uses one examination and five runs, and the observed variability may depend on decoding and task conditions.

\section{Native Persian Claim-Level Dataset}

Transferring LUH to Persian requires supervision aligned with both Persian responses and the target tokenizer. Rather than translating an English dataset and then re-establishing claim and token alignment, we construct the datasets directly in Persian. The same questions are answered by both backbones, allowing the resulting datasets to be paired.

\subsection{Dataset Construction}

The pipeline has five steps. First, we curate 1,600 Persian medical questions using a deterministic procedure. At most one question is kept per base medical entity to reduce near-duplicates; synthetic entity variants are capped at about 6\%; and questions are selected at even intervals across an alphabetically ordered pool of medical entities. These restrictions are intended to reduce template artifacts while preserving broad topical coverage.

Second, Gaokerena-V and Gaokerena-R answer the same 1,600 questions in the same order using greedy decoding. Thus, the two datasets are paired at the question level: differences in their responses and in the resulting heads arise from the backbones rather than from different question sets.

Third, atomic, verifiable claims are extracted directly from the Persian responses using Persian-language prompting. Sentences shorter than 40 characters after markup removal are discarded because they are treated as headings or labels rather than substantive factual assertions.

Fourth, each extracted claim is independently labeled by DeepSeek-V4-Flash~\cite{b13} as supported, hallucinated, or undecided. Undecided claims are retained in the released data for transparency but excluded from the supervised loss. The use of an automatic annotator makes the resource scalable, but it also means that the labels should not be treated as equivalent to a human-verified gold standard.

Fifth, each supported or hallucinated claim is mapped back to the exact token span in the Aya-Expanse tokenization of the response. Tokens inside the claim span receive the corresponding binary label, while tokens outside supervised claims are assigned the masking value $-100$. This alignment is performed against the target tokenizer rather than a translated representation.

Table~\ref{tab:data} summarizes the resulting paired datasets. The table reports not only the number of responses and extracted claims, but also the supported and hallucinated counts, undecided claims, hallucination rates, supervised-token coverage, and the positive-class weights used during training. These statistics make two properties of the resource explicit: the two backbones contribute similar hallucination prevalences despite producing different response lengths and claim counts, and the class imbalance is substantial enough to require weighted supervision.

\begin{table}[ht]

\centering

\small

\caption{Statistics of the two native Persian uncertainty datasets.}

\label{tab:data}

\resizebox{\columnwidth}{!}{%

\begin{tabular}{|l|c|c|}

\hline

& \textbf{LUH-V} & \textbf{LUH-R} \\ \hline

Annotated responses & 1,600 & 1,600 \\ \hline

Extracted claims & 60,434 & 54,209 \\ \hline

Supported claims & 48,991 & 43,305 \\ \hline

Hallucinated claims & 10,557 & 9,528 \\ \hline

Undecided claims & 886 & 1,376 \\ \hline

Hallucination rate & 17.73\% & 18.03\% \\ \hline

Supervised tokens & 61.4\% & 54.9\% \\ \hline

Positive class weight & 4.64 & 4.53 \\ \hline

\end{tabular}}

\end{table}

As shown in Table~\ref{tab:data}, the resulting datasets contain 60,434 and 54,209 extracted claims. Their overall hallucination rates are similar (17.73\% and 18.03\%), while the undecided counts differ because the two backbones generate different responses. The supervised-token percentages also differ, reflecting differences in response lengths and claim spans. The nearly matched hallucination prevalences are useful for the later backbone comparison because differences in PR-AUC are not driven simply by a large difference in the overall positive rate of the two datasets.

Each dataset is divided into training, validation, and test splits containing 1,458, 92, and 50 responses, respectively. The splits are sampled at even intervals across the ordered entity pool rather than taking contiguous blocks, so all three partitions cover the topical range.

\section{Uncertainty Head}

\subsection{Architecture}

We train a claim-level head over each frozen backbone. The head pools token-level signals associated with each claim and produces one hallucination score per claim, matching the granularity of the labels and the intended downstream use. We do not fine-tune the underlying language model.

Following the LUH design~\cite{b1}, the head receives two feature families. The first is the backbone's attention history from all layers, using an attention-history size of three preceding positions without pooling at the feature-extraction stage. The second consists of the top-4 token probabilities at each position. These features are concatenated and processed by a two-layer Transformer encoder with hidden dimension 768, eight attention heads, and dropout 0.1. The resulting representation is mapped to a claim-level hallucination prediction.

This design preserves the generator's original parameters and uses information already computed during generation. Consequently, inference requires one normal forward pass followed by the auxiliary prediction rather than a second generation, external retrieval step, or repeated sampling.

\subsection{Training Objective}

For claim $c$, let $y_c=1$ denote a hallucinated claim, $y_c=0$ a supported claim, and $p_c$ the predicted hallucination probability. With $\mathcal{C}$ denoting the labeled claims in a batch, we use weighted binary cross-entropy:

\[
\mathcal{L} =
-\frac{1}{|\mathcal{C}|}
\sum_{c\in\mathcal{C}}
\left[
 w_{+} y_c \log p_c + (1-y_c)\log(1-p_c)
\right],
\]

where $w_{+}$ is 4.64 for Gaokerena-V and 4.53 for Gaokerena-R. The weights compensate for the approximately four-to-one imbalance between supported and hallucinated claims.

Training uses the same optimization configuration for both backbones. The complete configuration is shown in Table~\ref{tab:training}. We do not perform hyperparameter optimization; checkpoint selection is based on validation PR-AUC.

\begin{table}[ht]

\centering

\small

\caption{Training configuration used for both uncertainty heads.}

\label{tab:training}

\resizebox{\columnwidth}{!}{%

\begin{tabular}{|l|c|}

\hline

\textbf{Training setting} & \textbf{Value} \\ \hline

Maximum epochs & 10 \\ \hline

Learning rate & $1\times10^{-4}$ \\ \hline

Warmup ratio & 0.05 \\ \hline

Weight decay & 0.1 \\ \hline

Per-device batch size & 8 \\ \hline

Gradient accumulation & 1 \\ \hline

Maximum gradient norm & 1.0 \\ \hline

Head dimension / layers / heads & 768 / 2 / 8 \\ \hline

Head dropout & 0.1 \\ \hline

Backbone precision & \texttt{fp16} \\ \hline

Checkpoint metric & PR-AUC \\ \hline

Early stopping & 3 epochs \\ \hline

\end{tabular}}

\end{table}

\subsection{Adapter Merging and Reproducibility}

The Gaokerena-V and Gaokerena-R checkpoints are distributed as LoRA adapters rather than standalone language-model weights. The public LUH trainer loads a backbone through a standard model constructor and does not directly handle the Gaokerena PEFT adapter in this path. We therefore load Aya-Expanse-8B, apply each adapter with PEFT, merge the adapter into the base model, and train the uncertainty head against the resulting standalone model.

Both notebooks use half-precision loading for the merged model and the training configuration. Training is performed on a single CUDA GPU with approximately 40~GB of memory at batch size 8. The backbone parameters remain frozen throughout training; only parameters belonging to the uncertainty head are updated.

We also verify tokenizer compatibility before training. The notebooks compare the adapter and base tokenizers and check that chat-template tokenization reproduces the stored input IDs for the test split. This guards against a failure mode in which the supervision refers to token positions generated by a different tokenizer.

\subsection{Implementation Corrections}

The public LUH code contained two implementation issues that we corrected before training. First, under recent versions of Transformers, tokenized output from \texttt{apply\_chat\_template} is a mapping rather than a plain list. The original code used its length as a token count, which can shift the computed reply boundary and place claim masks on prompt positions. We explicitly extract the \texttt{input\_ids} sequence before computing the boundary.

Second, the original model-loading code did not correctly resolve the configured PyTorch data type. As a result, a configuration requesting \texttt{fp16} could still load the 8B backbone in \texttt{fp32}. We replace this resolution with an explicit lookup of the requested torch dtype. This makes the configured precision effective and keeps memory usage within the intended budget.

These corrections affect preprocessing and model loading rather than the uncertainty-head architecture. We document them because silent masking or precision errors can otherwise produce apparently valid training runs with incorrect supervision or unnecessary memory usage.

\subsection{Validation and Test Isolation}

The LUH training framework uses one named validation split internally for evaluation, checkpoint selection, early stopping, and prediction. We therefore configure the 92-response \texttt{eval} split as the validation set during training and do not expose the 50-response \texttt{test} split to checkpoint selection.

After training, a separate evaluation-only run loads the selected head and changes the framework's validation argument to \texttt{test}. The final test metrics therefore do not influence training or model selection. This two-stage procedure avoids using the final test responses as an implicit tuning set.

\section{Results}

Table~\ref{tab:results} reports claim-level performance on the held-out test splits. Hallucinated claims constitute 20.95\% of the 1,951 Gaokerena-V test claims and 17.46\% of the 1,687 Gaokerena-R test claims. A classifier that never flags a claim would consequently achieve 79.05\% and 82.54\% accuracy, so accuracy alone is not informative for comparing the two detectors. The expected PR-AUC of a random scorer equals the positive base rate, giving random baselines of 0.2095 and 0.1746.

\begin{table}[ht]

\centering

\small

\caption{Claim-level uncertainty-head performance on the held-out test splits (\%).}

\label{tab:results}

\resizebox{\columnwidth}{!}{%

\begin{tabular}{|l|c|c|}

\hline

\textbf{Metric} & \textbf{Gaokerena-V} & \textbf{Gaokerena-R} \\ \hline

Accuracy & 76.24 & 62.96 \\ \hline

Precision & 44.85 & 29.52 \\ \hline

Recall & 58.17 & 80.84 \\ \hline

F1 & 50.65 & 43.24 \\ \hline

ROC-AUC & 78.52 & 78.10 \\ \hline

PR-AUC & 48.20 & 46.52 \\ \hline

Random PR-AUC & 20.95 & 17.46 \\
 (base rate) & &  \\ \hline

PR-AUC / random & 2.30 & 2.66 \\ \hline

Test claims & 1,951 & 1,687 \\ \hline

Selected epoch & 3 & 4 \\ \hline

\end{tabular}}

\end{table}

\subsection{Gaokerena-V}

The Gaokerena-V head reaches a PR-AUC of 0.4820 and a ROC-AUC of 0.7852. At the selected operating point, it recalls 58.17\% of hallucinated claims with 44.85\% precision. Its validation PR-AUC reaches 0.4911 at the selected checkpoint, and the final test PR-AUC is 0.4820. The selected checkpoint corresponds to epoch 3, with no later checkpoint improving the validation criterion.

The difference between the random PR-AUC baseline (0.2095) and the observed PR-AUC indicates that the head captures information associated with hallucination labels in the frozen backbone features. This signal is obtained without retrieval, reference passages, or additional generations.

\subsection{Gaokerena-R}

The Gaokerena-R head reaches a PR-AUC of 0.4652 and a ROC-AUC of 0.7810. Its selected operating point has higher recall (80.84\%) and lower precision (29.52\%) than Gaokerena-V. The validation PR-AUC at the selected checkpoint is 0.4108, and the selected checkpoint is epoch 4, followed by early stopping after three epochs without validation improvement.

The lower raw PR-AUC relative to Gaokerena-V should be interpreted alongside the different positive base rates. Relative to the corresponding random floors, the Gaokerena-R and Gaokerena-V heads reach 2.66 and 2.30 times random performance, respectively.

\subsection{Comparison Across Backbones}

The ROC-AUC values are close (0.7852 for Gaokerena-V and 0.7810 for Gaokerena-R), while their precision-recall operating points differ considerably. On these held-out splits, both backbones therefore provide internal signals from which the same LUH-style architecture can learn useful hallucination ranking.

This comparison is enabled by the paired dataset design: both backbones answer the same 1,600 questions, use the same claim-extraction and verification pipeline, and are trained with the same head architecture and optimization settings. The comparison should nevertheless be interpreted as evidence from two backbones within the same Persian medical setting, rather than as evidence that language models generally expose equally useful internal uncertainty signals.

\subsection{Comparison with the Original UHead Results}

The comparison with the original UHead study~\cite{b1} is important because it tests whether the LUH-style design remains useful after changing both the backbone family and the language. The original study reports a PR-AUC of 0.66 for its Mistral 7B in-domain biographies setting and 0.34--0.49 across seven out-of-domain English domains. It also reports 0.44--0.58 when a Gemma 2 9B uncertainty head trained on English biographies is evaluated on Russian, Chinese, and German data, in addition to English in-domain data. Our Gaokerena-V and Gaokerena-R values of 0.4820 and 0.4652 therefore fall inside both the original study's reported out-of-domain range for Mistral and its cross-lingual range for Gemma. In particular, the two Persian results are close to the upper end of the original Mistral OOD range and comfortably within the Gemma cross-lingual range.

A prevalence-normalized comparison leads to a similar conclusion. In the original Mistral experiments, the OOD UHead results correspond to approximately 2.15--4.70 times their respective random PR-AUC baselines, while the Gemma cross-lingual results correspond to approximately 1.71--3.54 times random. Our heads achieve 2.30 and 2.66 times random for Gaokerena-V and Gaokerena-R, respectively. Thus, the Persian heads retain a gain over chance that is of the same order as the original study across both its OOD and cross-lingual settings. This relative comparison is especially useful here because the positive-claim prevalence differs substantially across datasets, making raw PR-AUC alone difficult to interpret.

The feature choice also follows evidence from the original study. Among the feature sets evaluated there on Mistral 7B, the combination of attention and token probabilities achieved a validation PR-AUC of 0.642, outperforming variants based only on hidden states or alternative feature sets. Our head therefore retains the same general feature families while adapting the representation and supervision to the Aya-Expanse-derived Persian medical backbones. The resulting performance suggests that this feature combination can remain informative outside the original English and model settings, although the present experiments do not isolate the contribution of each feature family.

Table~\ref{tab:backbone_comparison} summarizes the comparison. The table is not intended as a direct leaderboard because the studies differ in backbone architecture, language, domain, dataset construction, claim prevalence, and evaluation protocol. Instead, it answers a narrower adaptation question: whether our single-pass heads produce PR-AUC and prevalence-normalized gains that are broadly consistent with the scale reported for the original UHead framework. On that criterion, both Persian heads fall within the empirical range established by the original work, supporting the view that the framework transfers meaningfully to a new low-resource medical setting.

\begin{table}[ht]

\centering

\scriptsize

\caption{Comparison with representative PR-AUC results from the original UHead study~\cite{b1}. Random-normalized values are computed as PR-AUC divided by the corresponding random baseline. Ranges summarize the heterogeneous sets reported in that study.}

\label{tab:backbone_comparison}

\resizebox{\columnwidth}{!}{%

\begin{tabular}{lccc}

\toprule

\textbf{Setting} & \textbf{PR-AUC} & \textbf{Random} & \textbf{PR-AUC / random} \\

\midrule

Mistral 7B, in-domain biographies & 0.66 & 0.29 & 2.28$\times$ \\

Mistral 7B, OOD domains & 0.34--0.49 & 0.11--0.26 & 2.15--4.70$\times$ \\

Gemma 2 9B, cross-lingual & 0.44--0.58 & 0.15--0.34 & 1.71--3.54$\times$ \\

\midrule

Gaokerena-V (ours) & 0.4820 & 0.2095 & 2.30$\times$ \\

Gaokerena-R (ours) & 0.4652 & 0.1746 & 2.66$\times$ \\

\bottomrule

\end{tabular}}

\end{table}

\subsection{Review Workload at the Selected Thresholds}

The reported precision, recall, and base rates can be translated into an approximate review workload for the selected checkpoints. On the Gaokerena-V test split, about 530 of 1,951 claims are flagged, corresponding to roughly 27\% of all claims, and about 238 of the flagged claims are hallucinations. On the Gaokerena-R split, about 807 of 1,687 claims are flagged, or roughly 48\% of all claims, again identifying about 238 hallucinated claims. These figures are derived from the reported operating-point metrics and are intended only to illustrate the precision-recall trade-off.

The two workloads illustrate why threshold selection is application-dependent. A reviewer who can inspect only a small fraction of claims would require a stricter operating threshold, while a reviewer prioritizing hallucination recall could accept more false positives. The present experiments use the default threshold of each selected checkpoint rather than tuning the threshold for a specific review budget.

\section{Conclusion}

We adapted the LLM Uncertainty Head framework to Aya-Expanse-based Persian medical models and constructed two paired Persian claim-level hallucination datasets. The resulting heads use frozen-backbone attention and token-probability features to produce claim-level hallucination scores without retrieval or repeated sampling. Across Gaokerena-V and Gaokerena-R, the heads obtain ROC-AUCs of 0.7852 and 0.7810 and PR-AUCs that are 2.30 and 2.66 times the corresponding random baselines. The paired design further enables a controlled comparison of the same uncertainty-head architecture across two differently trained Persian medical backbones. The comparison with the original UHead study shows that both Persian heads achieve PR-AUCs and prevalence-normalized gains within the range reported for the original framework across out-of-domain and cross-lingual evaluations, supporting the feasibility of adapting single-pass uncertainty heads to a new low-resource medical setting. Together, these results provide a reproducible starting point for single-pass uncertainty estimation in Persian medical language models.

\end{document}